\documentclass[letterpaper]{article} 
\usepackage{aaai2026}  
\usepackage{times}  
\usepackage{helvet}  
\usepackage{courier}  
\usepackage[hyphens]{url}  
\usepackage{graphicx} 
\usepackage{natbib}  
\usepackage{caption} 
\usepackage{algorithm}
\usepackage{algorithmic}

\usepackage{booktabs}
\usepackage{colortbl}
\usepackage{multirow}
\usepackage{makecell}
\usepackage{amsmath}
\usepackage{amsfonts}
\usepackage{subfig}
\usepackage{xcolor}
\usepackage{newfloat}
\usepackage{listings}
\DeclareCaptionStyle{ruled}{labelfont=normalfont,labelsep=colon,strut=off} 
\floatstyle{ruled}
\newfloat{listing}{tb}{lst}{}
\floatname{listing}{Listing}
\title{Distilling Cross-Modal Knowledge via Feature Disentanglement}
\author {
    Junhong Liu\textsuperscript{\rm 1},
    Yuan Zhang\textsuperscript{\rm 2},
    Tao Huang\textsuperscript{\rm 3},
    Wenchao Xu\textsuperscript{\rm 4},
    Renyu Yang\textsuperscript{\rm 1}\thanks{Corresponding Author}
}
\affiliations{
    \textsuperscript{\rm 1}School of Software, Beihang University \\
    \textsuperscript{\rm 2}School of Computer Science, Peking University\\
    \textsuperscript{\rm 3}Shanghai Jiao Tong University \\
    \textsuperscript{\rm 4}Hong Kong University of Science and Technology\\
    \{junhongliu, renyuyang\}@buaa.edu.cn, zhangyuan@alumni.pku.edu.cn, t.huang@sjtu.edu.cn, wenchaoxu@ust.hk
}

\usepackage{bibentry}

\begin{document}

\maketitle

\begin{abstract}
Knowledge distillation (KD) has proven highly effective for compressing large models and enhancing the performance of smaller ones. However, its effectiveness diminishes in cross-modal scenarios, such as vision-to-language distillation, where inconsistencies in representation across modalities lead to difficult knowledge transfer. To address this challenge, we propose frequency-decoupled cross-modal knowledge distillation, a method designed to decouple and balance knowledge transfer across modalities by leveraging frequency-domain features. We observed that low-frequency features exhibit high consistency across different modalities, whereas high-frequency features demonstrate extremely low cross-modal similarity. Accordingly, we apply distinct losses to these features: enforcing strong alignment in the low-frequency domain and introducing relaxed alignment for high-frequency features. We also propose a scale consistency loss to address distributional shifts between modalities, and employ a shared classifier to unify feature spaces. Extensive experiments across multiple benchmark datasets show our method substantially outperforms traditional KD and state-of-the-art cross-modal KD approaches.
\end{abstract}

\begin{links}
    \link{Code}{https://github.com/Johumliu/FD-CMKD}
\end{links}
\section{Introduction}
\label{sec:intro}

Knowledge Distillation (KD) has emerged as a fundamental technique for model compression and performance improvement. The core concept of KD involves utilizing a large and high-capacity teacher model to mentor a smaller yet more efficient student model. Through this process, the student model learns to approximate the behavior of the teacher model, often achieving comparable or even superior performance despite its reduced complexity.

Despite the substantial success of traditional distillation methods in unimodal settings \cite{huang2022masked, zhang2023avatar}, such as image or text tasks, many real-world applications inherently involve multimodal data, including vision, language, and audio. In these cross-modal scenarios, effectively transferring knowledge among modalities presents unique challenges. As a result, researchers have increasingly turned their attention to devise cross-modal knowledge distillation (CMKD) framework to enhance the performance of a student model in one modality by leveraging the knowledge of a teacher model in a different modality.

CMKD presents more formidable challenges than its single-modal KD. The root cause of this lies in the inherent representational inconsistency among features from different modalities. Specifically, features from each modality concurrently encode both the cross-modal shared semantic content (the ``\textit{what}") and the modality-specific detailed characteristics (the ``\textit{how}"). Consequently, directly imposing a unified, strong alignment loss on these heterogeneous features often leads to ``representational conflicts", compelling the student model to distort the unique characteristics of its native modality, thereby undermining its intrinsic expressive capabilities and impeding effective knowledge transfer.

While preliminary progress has been made in cross-modal distillation, existing methods \cite{gupta2016cross, thoker2019cross, afouras2020asr, liu2023emotionkd, cao2025move} often suffer from several limitations: they are typically restricted to specific scenarios or primarily focus on distillation from a stronger modality to a weaker one. Recently, C2KD \citep{huo2024c2kd} introduced a cross-modal distillation technique based on logits to reduce the gap between modalities. However, it overlooks the distillation of challenging samples with misaligned soft labels, which are crucial for effective cross-modal knowledge transfer. Furthermore, C2KD exclusively emphasizes logit-level distillation, while intermediate features, which encapsulate richer modal details and semantic information in cross-modal settings, play a more pivotal role in facilitating complementary and effective knowledge transfer. This motivates us to explore cross-modal feature distillation, addressing the challenges posed by discrepant feature representations and misaligned feature spaces.

\definecolor{myblue}{RGB}{30, 100, 200}
\definecolor{myred}{RGB}{200, 50, 50}

\begin{table}[t]
    \centering
    \setlength{\tabcolsep}{8pt} 
    \renewcommand{\arraystretch}{1.4} 
    \small
    
    \begin{tabular}{p{2.4cm}|p{1.8cm}|p{1.8cm}}
        \toprule
        \makecell[c]{\textbf{Feature}} & \makecell[c]{\textbf{CREMA-D}} & \makecell[c]{\textbf{AVE}} \\
        \hline
        \makecell[c]{\textbf{{Original}} Features} & \makecell[c]{$0.84$} & \makecell[c]{$0.74$} \\
        \hline
        \makecell[c]{\textbf{{Low}} Frequency} &\makecell[c]{$0.91$} & \makecell[c]{$0.85$} \\
        \hline
        \makecell[c]{\textbf{{High}} Frequency} & \makecell[c]{$-0.02$} & \makecell[c]{$-0.01$} \\
        \bottomrule
    \end{tabular}
    \caption{Cross-Modal Similarity Results for CREMA-D and AVE Datasets across Different Features.}
    \label{tab:motivation}
\end{table} 

Inspired by  \citep{williams2018wavelet,xu2020learning,pham2024frequency,zhang2024freekd}, we investigate the frequency representations of multimodal features, and show that features decomposed into different frequency bands exhibit varying levels of effectiveness for representation. Specifically, low-frequency features tend to encode more modality-shared semantic information, whereas high-frequency features exhibit greater modality-specificity. As illustrated in Table \ref{tab:motivation}, low-frequency features consistently demonstrate higher inter-modal similarity compared to raw features across various datasets. This suggests that low-frequency features serve as effective carriers of modality-agnostic semantic information. In contrast, high-frequency features exhibit remarkably low inter-modal similarity, which further substantiates that high-frequency features predominantly encode modality-specific details that are challenging to directly align. This provides critical insights into how to separate and independently process these two distinct types of information, thereby enabling the formulation of effective distillation strategies.

Based on these insights, we propose a new approach of decomposing features into low-frequency and high-frequency components for distillation, applying distinct loss functions to each of them, respectively. For low-frequency features, we employ the traditional mean squared error (MSE) loss to ensure ``strong consistency" such that modality-generic information from the teacher model can be better captured. Meanwhile, since high-frequency features contain modality-specific knowledge and exhibit greater variation, making the full alignment less suited, we introduce the logarithmic mean squared error (logMSE) loss to maintain ``weak consistency".  Furthermore, given that distribution differences are critical for effective knowledge transfer \citep{pan2009survey,li2019delta,sun2016deep}, we propose a scale consistency loss by the alignment of different modalities through feature standardization, to address the significant discrepancies between modalities. This allows the model to focus on intrinsic discriminative features and reduces the impact of scale variations. We also introduce a shared classifier to align feature spaces further to ensure consistent decision boundaries across modalities, enhancing the effectiveness of cross-modal distillation.

To summarize, our contributions are three-fold:

\begin{itemize}
    \item By analyzing features across different modalities, we found that low-frequency features exhibit stronger cross-modal similarity compared to high-frequency features. Based on this crucial insight, we propose a novel frequency-domain decoupled Cross-Modal Knowledge Distillation framework, specifically designed to address the inherent conflict between semantic and detailed information during cross-modal distillation.
    \item We designed a set of differentiated distillation strategies tailored to process the decoupled features. Furthermore, we introduce a scale consistency loss and employ a shared classifier to further optimize the alignment of cross-modal feature spaces.
    \item We perform extensive experiments on diverse datasets, covering various modalities and tasks and employing different network architectures to demonstrate the effectiveness of our approach. 
\end{itemize}
\section{Related Work}

\subsection{Generic Knowledge Distillation}
Traditional knowledge distillation methods fall into two main categories: \textbf{logit-based distillation} and \textbf{feature-based distillation}. Logit-based distillation, first introduced by \citep{DBLP:journals/corr/HintonVD15}, transfers knowledge by minimizing the Kullback-Leibler (KL) divergence between the teacher and student model outputs, helping the student learn inter-class relationships. DKD \citep{DBLP:conf/cvpr/ZhaoCSQL22} refines this process by separating target and non-target class distillation, allowing better learning of category-specific information. DML \citep{zhang2018deep} enhances transfer by having two models train each other as teachers, while DIST \citep{huang2022knowledge} uses correlation loss to improve logit distillation by capturing inter-class and intra-class relationships. Feature-based distillation uses intermediate features from the teacher model to supervise the student, aiding in better data representation. FitNet \citep{romero2014fitnets} was the first to have the student mimic the teacher’s intermediate features. Review \citep{chen2021distilling} introduced a mechanism that allows the student to learn teacher features layer by layer. Relational Knowledge Distillation (RKD) \citep{park2019relational} focuses on transferring relationships between samples, while PKD \citep{cao2022pkd} preserves relational information using Pearson correlation. FreeKD \citep{zhang2024freekd} starts to use frequency information to represent visual features for distillation on dense prediction tasks. OFD \citep{heo2019comprehensive} employs partial L2 loss, ignoring unhelpful features and focusing on beneficial ones. Generic Knowledge Distillation performs well in single-modal tasks, but due to differences in modality representation, it underperforms in cross-modal scenarios.

\subsection{Cross-modal Knowledge Distillation}
In cross-modal knowledge distillation (CMKD) research, various studies have explored effective methods for transferring knowledge across modalities. \citep{gupta2016cross} proposed an early CMKD framework that transferred labeled supervision from RGB images to depth and optical flow, enhancing the performance of these unlabeled modalities. For action recognition, studies such as \citep{thoker2019cross,dai2021learning,zhang2025autov,lee2023decomposed} leveraged RGB or optical flow to design CMKD frameworks that improved action detection accuracy. In medical image segmentation, \citep{wang2023learnable} addressed missing modalities by selecting the most contributive one for cross-modal distillation in multi-modal learning. CMKD has also been applied to tasks like camera-radar object detection and visual place recognition, as seen in works like \citep{zhao2024crkd,zhang2024unveiling,wang2024distilvpr}. These works are limited to specific scenarios or focus on distillation for individual modalities. \citep{xue2022modality} introduced the Modality Focusing Hypothesis (MFH), offering the first theoretical analysis of CMKD's effectiveness, highlighting modality-generic decisive features as crucial for knowledge transfer. More recently, \citep{huo2024c2kd} identified modality imbalance and soft label misalignment as major challenges for CMKD, and introduced the C2KD framework, which significantly improved performance through bidirectional distillation and dynamic selection. The works fall short in addressing the inconsistencies in specific modality information and fail to fully leverage modality-generic features for effective cross-modal transfer. Our work overcomes these limitations by introducing a frequency-domain feature decoupling approach.
\section{How Cross-Modal Features Differ?}

The major difference between conventional KD in single modality and cross-modal knowledge distillation (CMKD) is that, CMKD is designed to distill the knowledge from another different modality. This difference poses a significant challenge since the teacher and student are trained with data in different modalities, and therefore have more distinct feature representations. Therefore, to design our cross-modal feature distillation method, it is necessary to first analyze the difference of cross-modal features. In this section, we present two of our major findings: (i) modality-generic and modality-specific features act differently in frequency domain; (ii) The features of different modalities exhibit significant differences in scale.

\subsection{Decoupling Modality-specific and Modality-generic Features}
\label{sec:modality}

Feature vectors learned by deep neural networks are not arbitrary arrangements, but rather encode learned, structurally meaningful information. Features originating from a single modality typically encode both modality-shared generic semantic information and modality-specific idiosyncratic details concurrently~\citep{ngiam2011multimodal}. Therefore, for improved knowledge transfer in Cross-Modal Knowledge Distillation (CMKD), the effective disentanglement of these two types of features is paramount.

Inspired by frequency-domain decomposition principles in signal processing~\citep{bruna2013invariant}, we posit that the frequency composition of deep feature vectors naturally aligns with this semantic-detail hierarchy. Specifically, low-frequency components tend to capture smooth, macroscopic trends and patterns within feature vectors, which we hypothesize correspond to universal cross-modal semantic representations. Conversely, high-frequency components capture abrupt, fine-grained variations across feature vector dimensions, which we hypothesize primarily stem from modality-specific details or noise.

To validate the aforementioned hypothesis, we performed frequency-domain decomposition on feature vectors extracted from various modalities and quantified the inter-modal similarity of their resulting low- and high-frequency features. Cosine similarity was employed as the quantitative measure. As depicted in Table \ref{tab:motivation}, low-frequency features exhibited significantly high inter-modal similarity, even surpassing that of the raw features, whereas high-frequency features displayed extremely low inter-modal similarity. These results further corroborate our hypothesis and strongly suggest that for effective cross-modal knowledge distillation, features must be decomposed into low- and high-frequency components, which should then be processed differentially based on their distinct representational characteristics. We will formally present our proposed strategy in following section.

\begin{figure}[t]
  \centering
   \includegraphics[width=0.95\linewidth]{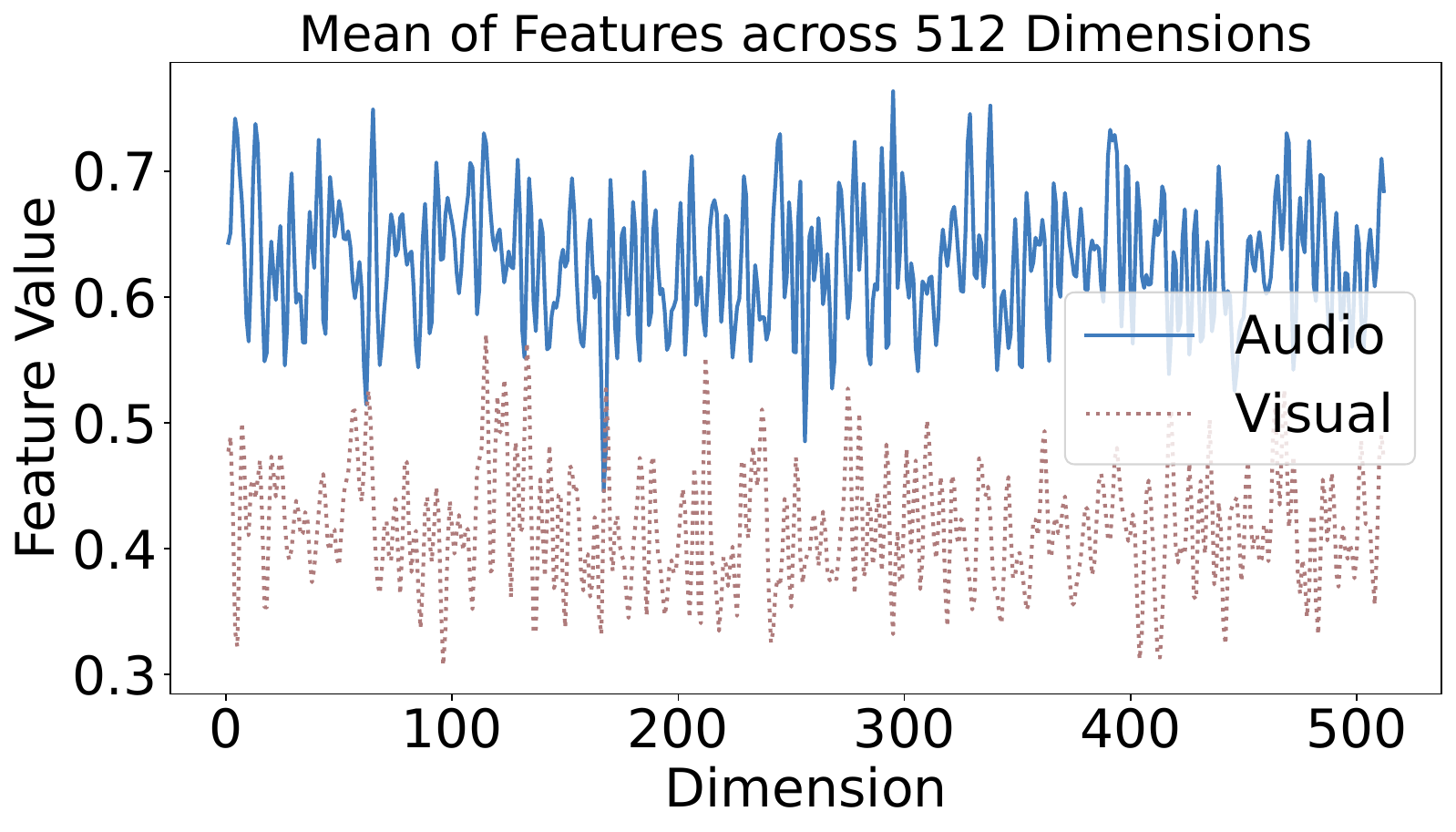}
   \caption{The comparison of feature mean value differences across modalities.}
   \label{fig:right_image}
\end{figure}

\subsection{Scale Differences Across Modalities}

In transfer learning, distribution differences play a crucial role in determining whether knowledge can be effectively transferred. If there are significant differences between the distributions of the source and target domains, the model may fail to capture useful information during the transfer process, leading to a substantial decrease in the effectiveness of the transfer.

We contend that inter-modal scale discrepancies constitute another contributing factor to suboptimal performance in cross-modal distillation. Through visualizing the feature means across 512 dimensions, it was revealed that significant scale differences exist among modalities. As shown in Figure \ref{fig:right_image}, the feature means of the audio and visual modalities differ across dimensions, with the feature values of the audio modality being noticeably higher than those of the visual modality.

When MSE is used to force the alignment of the student model's features with those of the teacher model, the student’s features may shift towards the mean of the teacher model’s features. However, this may conflict with the optimal mean expected in the student’s modality, leading to suboptimal performance in the student model.

Therefore, we should not directly use MSE loss to learn features from different modalities. Instead, we should design a loss function that respects the inherent scale differences between modalities to achieve more effective knowledge transfer.
\begin{figure*}[t]
    \centering
    \includegraphics[width=\textwidth]{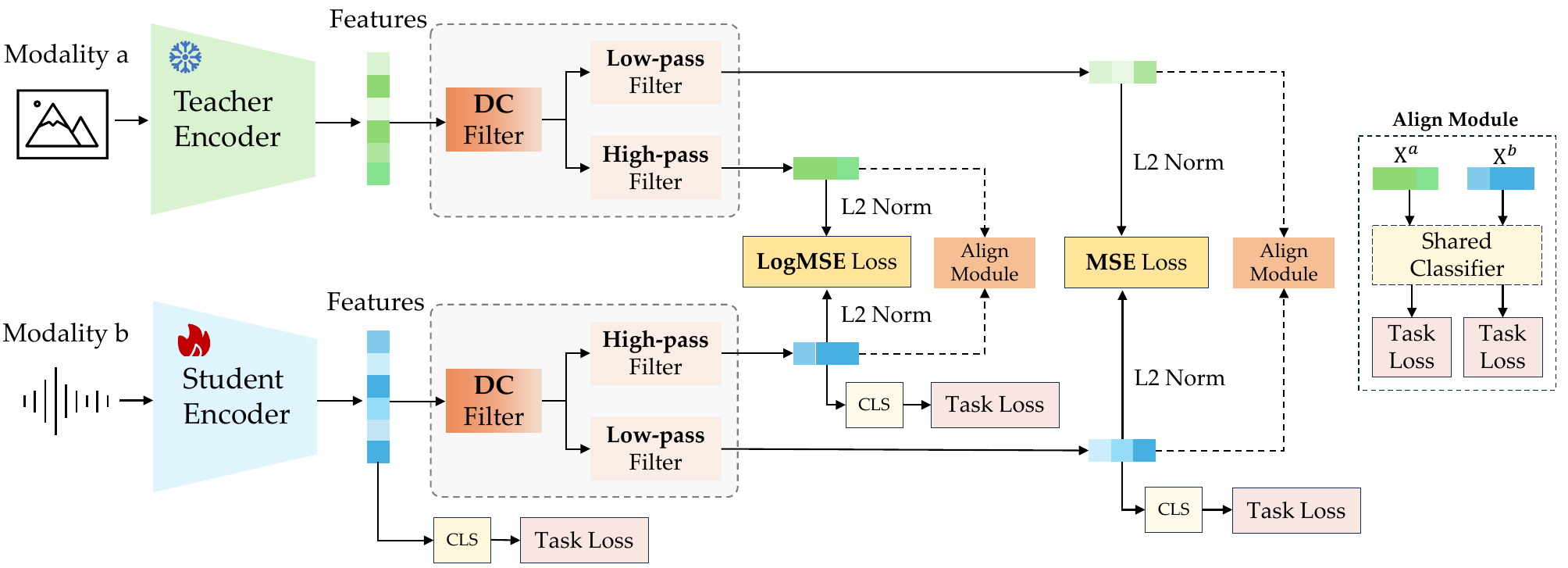}
    \caption{Framework of our method. We decouple the features of different modalities in the frequency domain into high-frequency and low-frequency components. For low-frequency features, MSE loss is applied, while logMSE loss is used for high-frequency features. Additionally, we ensure consistency in feature scale and feature space across modalities through feature normalization and alignment modules.}
    \label{fig:method}
\end{figure*}

\section{Our Approach}
\label{sec:method}

As previously discussed, we found that modality-specific and modality-generic information can be effectively decoupled through frequency domain analysis, and there are significant differences in the feature distributions across different modalities. In this section, we will formally introduce our method to improve CMKD: (i) We decouple the features into low-frequency and high-frequency components and apply different loss functions for distillation accordingly. (ii) We ensure the features from different modalities are consistent in scale and feature space. 

\subsection{Frequency-Decoupled Distillation}

We identified frequency decoupling of features as a effective way to disentangle the modality-generic and modality-specific information in the features. Formally, given the original feature $\mathbf{X}^m \in \mathbb{R}^{D}$ for a certain modality $m$, we compose the following three computation steps to decouple it into two features, namely low-frequency feature $\mathbf{X}_{\text{low}}^m$ and high-frequency feature $\mathbf{X}_{\text{high}}^m$.

\noindent \textbf{Spatio-temporal domain to frequency domain.} To decouple the original features, we first use Fourier transform to convert them into frequency domain, \textit{i.e.},
\begin{equation}
\mathbf{X}_f^m = \mathbf{DFT}(\mathbf{X}^m),
\end{equation}
where $\mathbf{X}_f^m$ represents the corresponding Fourier-transformed feature in complex frequency domain.

\noindent \textbf{High-pass and low-pass filtering.} In the frequency domain, we decompose $\mathbf{X}_f^m$ into different frequency components by designing a low-pass filter $\mathbf{M}_\text{low}$ and a high-pass filter $\mathbf{M}_\text{high}$. $\mathbf{M}_\text{low}$ and $\mathbf{M}_\text{high}$ are fixed binary mask filters: $\mathbf{M}_\text{low}$ sets the first half of the frequency components to 1 (low-pass filter), and $\mathbf{M}_\text{high}$ sets the second half to 1 (high-pass filter). Then the low-frequency part $\mathbf{X}_{\text{low},f}^m$ and the high-frequency part $\mathbf{X}_{\text{high},f}^m$ are computed as follows:
\begin{equation}
\mathbf{X}_{f,\text{low}}^m = \mathbf{X}_f^m \cdot \mathbf{M}_\text{low}, \quad \mathbf{X}_{f,\text{high}}^m = \mathbf{X}_f^m \cdot \mathbf{M}_\text{high}.
\end{equation}

\noindent \textbf{Feature reconstruction with inverse Fourier transform.} To obtain the reconstructed low-frequency and high-frequency features, we apply the Inverse Discrete Fourier Transform (IDFT) to transform the low-frequency and high-frequency components from the frequency domain back to the spatio-temporal domain, then we can obtain the decoupled features as
\begin{equation}
\mathbf{X}_{\text{low}}^m = \mathbf{IDFT}(\mathbf{X}_{f,\text{low}}^m), \quad \mathbf{X}_{\text{high}}^m = \mathbf{IDFT}(\mathbf{X}_{f,\text{high}}^m).
\end{equation}

The next task is to pinpoint the most suitable design of distillation loss for each type of features, respectively. As analyzed in previous section, low-frequency features primarily encompass modality-generic information, highly shared across different modalities. Hence, it is imperative to  maintain ``strong consistency'' for low-frequency features across different modalities so that their generality can be guaranteed. On the other hand, high-frequency features tend to capture modality-specific fine-grained information and are often accompanied by more noises. To preserve modality-specific details whilst reducing sensitivity to large errors stemming from the noises, we only require ``weak consistency" for high-frequency features across different modalities.

As a result, for the low-frequency features on two different modalities $a$ and $b$, we use the conventional mean square error (MSE) as the loss function, \textit{i.e.},
\begin{equation}
\mathcal{L}_{\text{low}} = \frac{1}{ND} \left\| \mathbf{X}_{\text{low}}^a - \mathbf{X}_{\text{low}}^b \right\|^2,
\label{eq:4}
\end{equation}
where $N$ and $D$ denote the batch size and dimension, respectively.

\begin{figure}[t]
  \centering
   \includegraphics[width=0.95\linewidth]{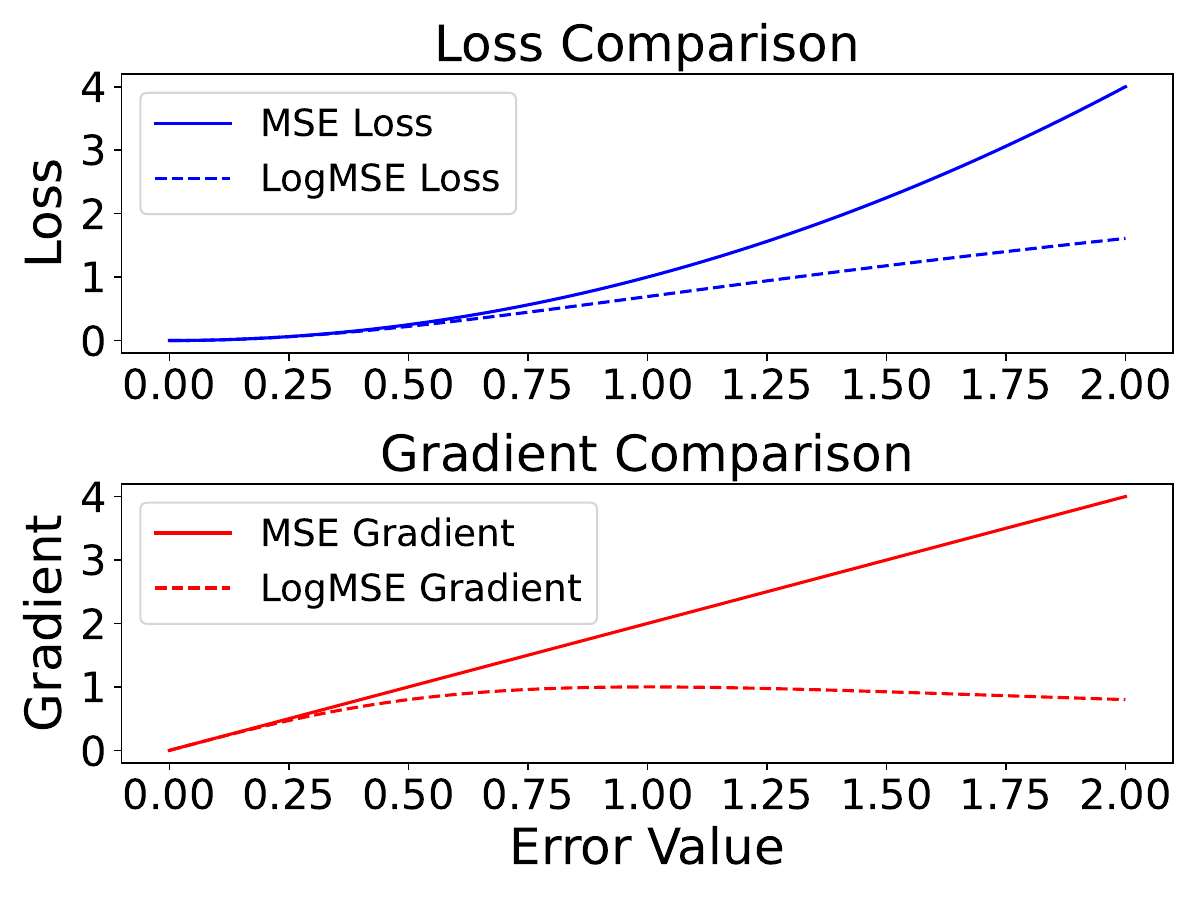}
   \caption{Comparison of value and gradient between MSE and logMSE losses.}
   \label{fig:logmse}
\end{figure}

While for the distillation of high-frequency features, a proper way is suppressing the significant gradient values caused by the noises and abnormally-large features. To this end, we leverage log mean square error (LogMSE) as the distillation loss, which has smoother gradients when the difference of two feature values is large, as shown in Figure \ref{fig:logmse}. The distillation loss for high-frequency features is formulated as
\begin{equation}
\mathcal{L}_{\text{high}} = \frac{1}{ND} \left\| \mathbf{\sigma}(\mathbf{X}_{\text{high}}^a) - \mathbf{\sigma}(\mathbf{X}_{\text{high}}^b) \right\|^2
\label{eq:5}
\end{equation}
\begin{equation}
\text{with}\quad\mathbf{\sigma}(\mathbf{X}) =
\begin{cases} 
\mathbf{log}(1 + \mathbf{X}), &  \mathbf{X} \geq 0 \\
-\mathbf{log}(1 - \mathbf{X}), & \mathbf{X} < 0
\end{cases},
\end{equation}
where $N$ and $D$ denotes the batch size and dimension, respectively.

\subsection{Alignment of Feature Scale and Feature Space}

The consistency of feature distributions is pivotal for knowledge transfer. However, the significant differences in feature distributions between different modalities result in poor performance in cross-modal knowledge transfer. To mitigate the distribution discrepancies between modalities, we propose solutions from both the feature scale and feature space perspectives.

\noindent \textbf{Feature scale alignment.} The inconsistency in feature scales is typically reflected in the fact that feature vectors from different modalities may have varying numerical ranges, which can negatively impact the effectiveness of knowledge distillation. To achieve feature scale alignment, we employed a Feature Standardization strategy, which includes the following steps:
\begin{enumerate}
\item Mean Subtraction: First, mean subtraction is applied to the feature vectors to ensure that the mean of the features is zero, eliminating any bias in the features.
\item L2 Normalization: Next, L2 normalization is performed on the zero-centered feature vectors to ensure that the L2 norm of each feature vector is 1. This ensures that all feature vectors are compared on the same scale, avoiding computational biases caused by differences in the lengths of the feature vectors.
\end{enumerate}

Herein is the formula for feature standardization:
\begin{equation}
\mathbf{Std}(\mathbf{X}) = \frac{\mathbf{X} - \bar{\mathbf{X}}}{\|\mathbf{X} - \bar{\mathbf{X}}\|_2},
\end{equation}
where $\mathbf{X}$ represents the input feature vector, $\bar{\mathbf{X}}$ represents the mean of the features, and $\|\cdot\|_2$ denotes the L2 norm. In practice, the mean subtraction operation can be directly implemented by using a DC filter in the frequency domain (as shown in Figure \ref{fig:method}). By doing so, the previous distillation losses in  Eq. \ref{eq:4} and  Eq. \ref{eq:5} can be reformulated as follows:
\begin{align}
    \mathcal{L}_{\text{low}} &= \frac{1}{ND} \left\| \mathbf{Std}(\mathbf{X}_{\text{low}}^a) - \mathbf{Std}(\mathbf{X}_{\text{low}}^b) \right\|^2,\\
    \mathcal{L}_{\text{high}} &= \frac{1}{ND} \left\| \mathbf{\sigma}(\mathbf{Std}(\mathbf{X}_{\text{high}}^a)) - \mathbf{\sigma}(\mathbf{Std}(\mathbf{X}_{\text{high}}^b)) \right\|^2.
\end{align}
\noindent \textbf{Feature space alignment.} Although feature scale alignment can alleviate the inconsistency in the numerical ranges of features from different modalities, solely relying on scale alignment is insufficient to address the fundamental differences in feature distributions across modalities. Features from different modalities not only differ in numerical scales but may also exhibit significant variations in the specific shapes of their distributions and the delineation of class boundaries.

To further enhance the effective transfer of cross-modal knowledge, we propose an alignment strategy from the perspective of feature space, ensuring that the features of the teacher model and the student model are comparable within the same space, thereby narrowing the distribution differences between modalities.

As shown in Figure \ref{fig:method}, we designed an alignment module based on a shared classifier to achieve feature space alignment. Through the shared classifier, the features of both the teacher model and the student model can be aligned within the same decision space, thus reducing the distribution differences between modalities. Specifically, the features from the teacher model and the student model are fed into the same shared classifier, where they are classified through the shared classifier, and the classification alignment loss is defined as follows:
\begin{equation}
\begin{split}
\mathcal{L}_{\text{align}} = & \; \mathbf{CE}(\mathbf{\Phi}_h(\mathbf{X}_{\text{high}}^a), y) 
+ \mathbf{CE}(\mathbf{\Phi}_h(\mathbf{X}_{\text{high}}^b), y) \\
& + \mathbf{CE}({\Phi}_l(\mathbf{X}_{\text{low}}^a), y)  
+ \mathbf{CE}({\Phi}_l(\mathbf{X}_{\text{low}}^b), y),
\end{split}
\end{equation}
where $\mathbf{CE}(\cdot)$ denotes the cross-entropy loss, $\mathbf{\Phi}_h$ and $\mathbf{\Phi}_l$ represent the shared classifiers for high-frequency and low-frequency features, respectively, and $y$ denotes the ground truth labels.

\noindent \textbf{Overall loss function.} In addition to the aforementioned losses, we also compute the cross-entropy loss on the raw features, low-frequency features, and high-frequency features of the student model, respectively, to ensure that these features are discriminative \citep{mao2023cross}. We denote this loss as $\mathcal{L}_\text{task}$. See Figure~\ref{fig:method} for an easier reference of all the losses we conduct. As a result,  the total loss function can be expressed as follows:
\begin{equation}
\label{eq:10}
\mathcal{L}_{\text{total}} = \mathcal{L}_{\text{task}} + \mathcal{L}_{\text{align}} + \lambda_1 \mathcal{L}_{\text{low}} + \lambda_2 \mathcal{L}_{\text{high}},
\end{equation}
where $\lambda_1$ and $\lambda_2$ represent the weighting parameters for the distillation losses of low-frequency and high-frequency features, respectively.

\renewcommand{\multirowsetup}{\centering}

\begin{table*}[t]
    \centering
    \renewcommand{\arraystretch}{1}
    \small
    \centering
    
    \begin{tabular}{p{1.8cm}| p{1.2cm}|p{1cm} | p{1cm} |p{1cm} | p{1cm} | p{1cm} | p{1cm} | p{1cm} | p{1cm} }
        \toprule
        \multirow{2}*{\textbf{Category}} & \multirow{2}*{\textbf{Method}} & \multicolumn{2}{c|}{\textbf{CREMA-D}} & \multicolumn{2}{c|}{\textbf{AVE}} & \multicolumn{2}{c|}{\textbf{VGGSound}} & \multicolumn{2}{c}{\textbf{CrisisMMD}}\\
        \cline{3-10}

        ~ & ~ & \makecell[c]{A} & \makecell[c]{V} & \makecell[c]{A} & \makecell[c]{V} & \makecell[c]{A} & \makecell[c]{V} & \makecell[c]{T} & \makecell[c]{V} \\
        \hline
        Uni-Modal & w/o KD & \makecell[c]{62.4} & \makecell[c]{66.8} & \makecell[c]{63.7} & \makecell[c]{38.8}	& \makecell[c]{68.9} & \makecell[c]{44.9} & \makecell[c]{77.4} & \makecell[c]{70.2} \\
        \hline

        \multirow{5}*{Logits} & Logit & \makecell[c]{61.7} & \makecell[c]{62.6} & \makecell[c]{60.0} & \makecell[c]{39.1} & \makecell[c]{65.7} & \makecell[c]{45.4} & \makecell[c]{78.5} & \makecell[c]{70.5} \\

        ~ & DIST & \makecell[c]{62.2} & \makecell[c]{64.0} & \makecell[c]{62.4} & \makecell[c]{40.3} & \makecell[c]{66.4} & \makecell[c]{45.5} & \makecell[c]{78.3} & \makecell[c]{71.3} \\

        ~ & DML & \makecell[c]{52.7} & \makecell[c]{61.2} & \makecell[c]{60.2} & \makecell[c]{\underline{43.3}} & \makecell[c]{57.8} & \makecell[c]{44.9} & \makecell[c]{78.2} & \makecell[c]{71.2} \\

        ~ & NKD & \makecell[c]{\underline{62.4}} & \makecell[c]{61.8} & \makecell[c]{60.7} & \makecell[c]{38.1} & \makecell[c]{65.6} & \makecell[c]{44.9} & \makecell[c]{78.1} & \makecell[c]{71.2} \\

        ~ & DKD & \makecell[c]{61.0} & \makecell[c]{61.4} & \makecell[c]{60.5} & \makecell[c]{38.1} & \makecell[c]{64.4} & \makecell[c]{44.5} & \makecell[c]{\underline{79.0}} & \makecell[c]{70.7} \\

        \hline
        \multirow{4}*{Feature} & Feat & \makecell[c]{60.9} & \makecell[c]{64.3} & \makecell[c]{58.7} & \makecell[c]{39.6} & \makecell[c]{67.7} & \makecell[c]{45.5} & \makecell[c]{77.7} & \makecell[c]{70.8} \\

        ~ & PKD & \makecell[c]{60.4} & \makecell[c]{\underline{64.8}} & \makecell[c]{58.0} & \makecell[c]{41.0} & \makecell[c]{62.9} & \makecell[c]{46.9} & \makecell[c]{77.5} & \makecell[c]{70.9} \\

        ~ & OFD & \makecell[c]{60.6} & \makecell[c]{61.6} & \makecell[c]{58.0} & \makecell[c]{39.6} & \makecell[c]{68.5} & \makecell[c]{45.8} & \makecell[c]{78.1} & \makecell[c]{71.2} \\

        ~ & AFD & \makecell[c]{61.2} & \makecell[c]{59.5} & \makecell[c]{\underline{62.7}} & \makecell[c]{38.8} & \makecell[c]{\underline{68.7}} & \makecell[c]{45.8} & \makecell[c]{69.8} & \makecell[c]{\underline{72.3}} \\

        \hline
        \multirow{2}*{Cross-Modal} & C2KD & \makecell[c]{57.5} & \makecell[c]{59.8} & \makecell[c]{\underline{62.7}} & \makecell[c]{39.3} & \makecell[c]{67.0} & \makecell[c]{\underline{47.9}} & \makecell[c]{77.9} & \makecell[c]{71.4} \\
    
        ~ & Ours & \makecell[c]{\textbf{64.1}} & \makecell[c]{\textbf{71.0}} & \makecell[c]{\textbf{64.9}} & \makecell[c]{\textbf{47.8}} & \makecell[c]{\textbf{70.0}} & \makecell[c]{\textbf{48.1}} & \makecell[c]{\textbf{79.1}} & \makecell[c]{\textbf{72.7}} \\

        \bottomrule
    \end{tabular}
    \caption{The comparison of methods on Audio-Visual and Image-Text classification tasks. The metric is the top-1 accuracy(\%). `A’, `V’, and `T’ represent Audio, Visual, and Text modalities, respectively. ``Uni" refers to unimodal models without distillation. ``Logit" and ``Feat" correspond to the original logit-based and feature-based distillation methods. ``C2KD" represents the cross-modal distillation method mentioned in \citep{huo2024c2kd}. The best is in \textbf{bold}, and the second is \underline{underlined}.}
    \label{tab:main}
\end{table*}

\section{Experiments}
We evaluate our method on classification and semantic segmentation tasks across various multimodal datasets. We provide experimental settings before detailing the result analysis.

\subsection{Classification Task}
\label{sec:cls_task}

\renewcommand{\multirowsetup}{\centering}

\begin{table*}[t]
    \centering
    \renewcommand{\arraystretch}{1.1}
    \small
    \centering
    
    \begin{tabular}{p{1.2cm}| p{1cm} | p{1cm} |p{1cm} | p{1cm} | p{1cm} | p{1cm} | p{1cm} | p{1cm} | p{1cm}}
        \toprule
        \textbf{Method} & \makecell[c]{Uni} & \makecell[c]{Logit} & \makecell[c]{DIST} & \makecell[c]{DKD} & \makecell[c]{Feat} & \makecell[c]{PKD} & \makecell[c]{AFD} & \makecell[c]{C2KD} & \makecell[c]{Ours} \\

        \hline
        Depth & \makecell[c]{30.9} & \makecell[c]{29.7} & \makecell[c]{\underline{32.3}} & \makecell[c]{32.5} & \makecell[c]{29.4} & \makecell[c]{31.0} & \makecell[c]{30.2} & \makecell[c]{31.8} & \makecell[c]{\textbf{33.2}} \\

        RGB & \makecell[c]{34.1} & \makecell[c]{32.8} & \makecell[c]{34.9} & \makecell[c]{\underline{35.3}} & \makecell[c]{32.8} & \makecell[c]{33.7} & \makecell[c]{32.7} & \makecell[c]{34.8} & \makecell[c]{\textbf{36.9}} \\

        \bottomrule
    \end{tabular}
    \caption{The comparison on the semantic segmentation task. The metric denotes the mean Intersection over Union (mIoU).}
    \label{tab:seg}
\end{table*}

\textbf{Dataset.} CREMA-D \citep{cao2014crema} is an emotion recognition dataset with audio and vision, featuring six emotions: happy, sad, angry, fear, disgust, neutral.
AVE \citep{tian2018audio} is an audio-visual event localization dataset with $4,143$ videos across $28$ event categories.
While VGGSound \citep{chen2020vggsound} is a large-scale audio-visual dataset with $210$K ten-second videos, a subset of $50$ categories for our experiments.
CrisisMMD \citep{alam2018crisismmd} is a multimodal dataset for natural disaster research, including annotated tweets and images from Twitter in image and text formats. For more detailed information about the dataset, please refer to Appendix.

\noindent \textbf{Experimental Settings.} Our experimental settings follow \citep{huo2024c2kd,fan2024detached,wei2024enhancing}. We use the ResNet-18 \citep{he2016deep} as the backbone for audio-visual datasets and train them for $100$ epochs in total. In the CrisisMMD dataset, we employ BERT-base \citep{devlin2018bert} and MobileNetV2 \citep{sandler2018mobilenetv2} to extract text and visual features, respectively. We only train text modality for $20$ epochs. We utilize the SGD optimizer with a momentum of $0.9$, and the batch size for training is set to $64$. For detailed training information, see Appendix.

\noindent \textbf{Results Analysis.} In Table \ref{tab:main}, we present the performance of our method on classification benchmarks. We compare the logit-based, feature-based, and cross-modal state-of-the-art distillation methods. that our proposed method consistently achieves the best performance across all datasets and modalities. For example, on the AVE dataset's visual modality, our method improves performance by 9\%, reaching 47.8\%, compared to the unimodal baseline. This highlights the effectiveness of our approach in transferring knowledge across modalities. Notably, our method excels in transferring knowledge from low-performing modalities to high-performing ones, where other methods fail. For instance, on CREMA-D's visual modality, AVE's audio modality, and VGGSound's audio modality, most methods underperform compared to the unimodal baseline, while our approach consistently improves performance by effectively transferring knowledge from weaker modalities. Additionally, our method is stable in bidirectional cross-modal transfer. On CrisisMMD, while DKD works well for text but not visual, and AFD succeeds for visual but fails for text, our method performs consistently across both modalities, achieving 79.1\% on text and 72.7\% on visual. This outstanding performance is attributed to our method's ability to capture both modality-specific and modality-agnostic information through frequency decomposition and customized loss functions, as well as mitigating inherent feature distribution differences through feature alignment. This ensures robust results across various modality pairs (A-V, T-V) and network architectures (ResNet-ResNet, BERT-MobileNet).

\subsection{Semantic Segmentation Task}
\textbf{Dataset.} NYU-Depth V2 \citep{icra_2019_fastdepth} is a multimodal dataset for indoor scene understanding research. It provides two modalities of depth information and RGB image information. There are a total of $40$ categories. It contains $1,449$ densely labeled RGB and depth image alignment pairs.


\noindent \textbf{Experimental Settings.} Following C2KD \citep{huo2024c2kd}, the DeepLab V3+ \citep{chen2018encoder} model is utilized with ResNet-18 as the backbone, which is initialized with the pre-trained weights on ImageNet \citep{deng2009imagenet}. We train the student for $50$ epochs in total and the batch size is $16$.

\noindent \textbf{Results Analysis.} Regarding segmentation task, Table \ref{tab:seg} shows the performance of various KD methods on NYU-Depth V2. Our method still consistently outperforms all other methods, with 33.2\% mIoU for Depth and 36.9\% mIoU for RGB. These results surpass the next best method (DIST for Depth, DKD for RGB) by a notable margin of 0.9\% and 1.6\%, respectively. As highlighted earlier in the classification tasks, our method is also stable in bidirectional cross-modal transfer in segmentation tasks.

\section{Analysis}
In this section, we first evaluate the effectiveness of the key components in our CMKD method, including frequency decomposition, feature alignment, and loss functions, through ablation studies. Then, we visualize the feature distributions using t-SNE, showcasing the improved feature separation and cross-modal knowledge transfer achieved by our proposed method, when compared with traditional techniques. For further analysis, please refer to the Appendix.

\subsection{Effectiveness of components in CMKD}
\renewcommand{\multirowsetup}{\centering}

\begin{table}[t]
    \centering
    \renewcommand{\arraystretch}{1}
    \small
    \centering
    
    \begin{tabular}{p{0.58cm}| p{0.74cm}| p{0.74cm}| p{0.57cm}| p{0.57cm} | p{0.57cm} |p{0.57cm} | p{0.57cm}}
        \toprule
        \multicolumn{4}{c|}{\textbf{Method}} & \multicolumn{2}{c|}{\textbf{CREMAD}} & \multicolumn{2}{c}{\textbf{AVE}} \\
        \hline
        \makecell[c]{Freq} & \makecell[c]{Align} & \makecell[c]{Scale} & \makecell[c]{Log} & \makecell[c]{A} & \makecell[c]{V} & \makecell[c]{A} & \makecell[c]{V} \\ 
        \hline
        &  &  &  & \makecell[c]{60.9} & \makecell[c]{64.3} & \makecell[c]{58.7} & \makecell[c]{39.6} \\
        \makecell[c]{$\checkmark$} &  &  &  & \makecell[c]{60.8} & \makecell[c]{68.7} & \makecell[c]{61.0} & \makecell[c]{43.3} \\
        
        & \makecell[c]{$\checkmark$} &  &  & \makecell[c]{60.9} & \makecell[c]{67.9} & \makecell[c]{63.2} & \makecell[c]{41.3} \\

        \makecell[c]{$\checkmark$} & \makecell[c]{$\checkmark$} &  & & \makecell[c]{61.8} & \makecell[c]{68.7} & \makecell[c]{62.4} & \makecell[c]{45.8} \\
        
        \makecell[c]{$\checkmark$} & & \makecell[c]{$\checkmark$} &  & \makecell[c]{62.2} & \makecell[c]{70.0} & \makecell[c]{62.4} & \makecell[c]{44.8} \\

        \makecell[c]{$\checkmark$} & \makecell[c]{$\checkmark$} & \makecell[c]{$\checkmark$} &  & \makecell[c]{62.2} & \makecell[c]{70.6} & \makecell[c]{62.4} & \makecell[c]{46.0} \\

        \makecell[c]{$\checkmark$} & \makecell[c]{$\checkmark$} & \makecell[c]{$\checkmark$} & \makecell[c]{$\checkmark$} & \makecell[c]{64.1} & \makecell[c]{71.0} & \makecell[c]{64.9} & \makecell[c]{47.8} \\

        \bottomrule
    \end{tabular}
    \caption{Ablation study of our components. Freq: frequency decomposition; Align: feature space alignment; Scale: feature standardization; Log: logMSE loss on high-frequency features; Baseline: original feature distillation.}
    
    
    \label{tab:abl}

\end{table}
We perform experiments to show the effectiveness of each proposed component in CMKD in Table~\ref{tab:abl}. Firstly, it is evident that each individual component contributes positively to the overall performance. Frequency decomposition distillation provides improvements for most modalities as it helps to separate modality-specific information from modality-generic information. However, these improvements are not always consistent; for example, on the audio (A) modality of the CREMA-D dataset, there is a 0.1\% performance drop. This inconsistency may stem from significant differences in feature distributions across modalities. When we add the Feature space alignment and Feature standardization modules, the cross-modal performance improves significantly, highlighting the importance of reducing feature distribution discrepancies between modalities. Moreover, applying logMSE loss to high-frequency features enhances the transmission of modality-specific information, indicating that it is not necessary to fully align modality-specific information, and maintaining a weak consistency is more effective for transferring such information. Finally, the comprehensive integration of all components ensures more robust cross-modal knowledge transfer, thereby achieving more stable performance across different modalities.

\begin{figure}
    \centering
    \subfloat[Without KD]{\includegraphics[width=.33\columnwidth]{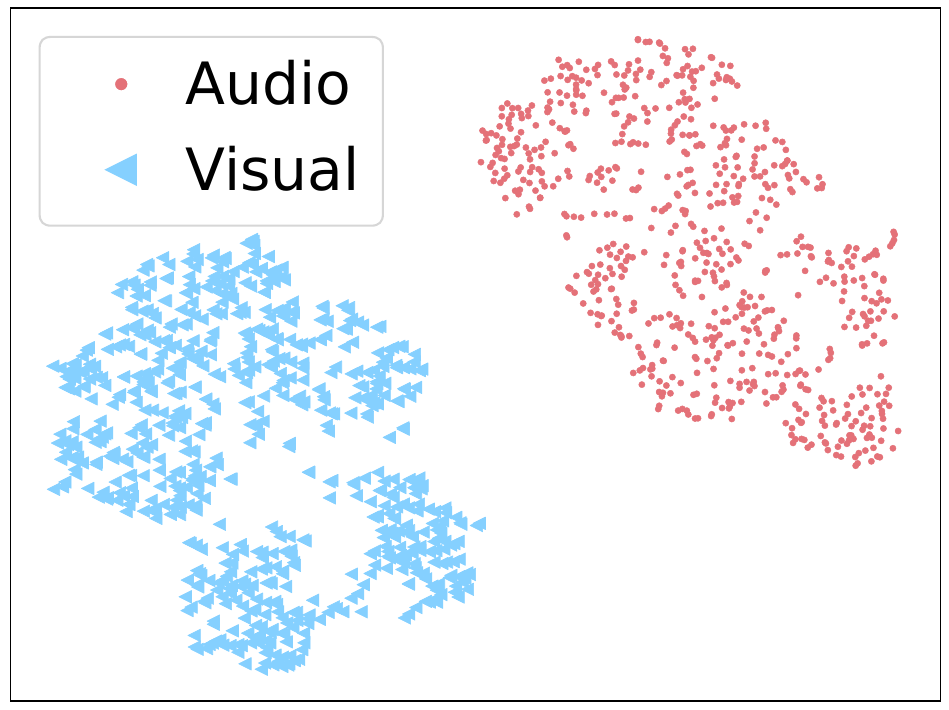}%
    \label{fig1:raw}}
    \hfil
    \subfloat[Feat KD]{\includegraphics[width=.33\columnwidth]{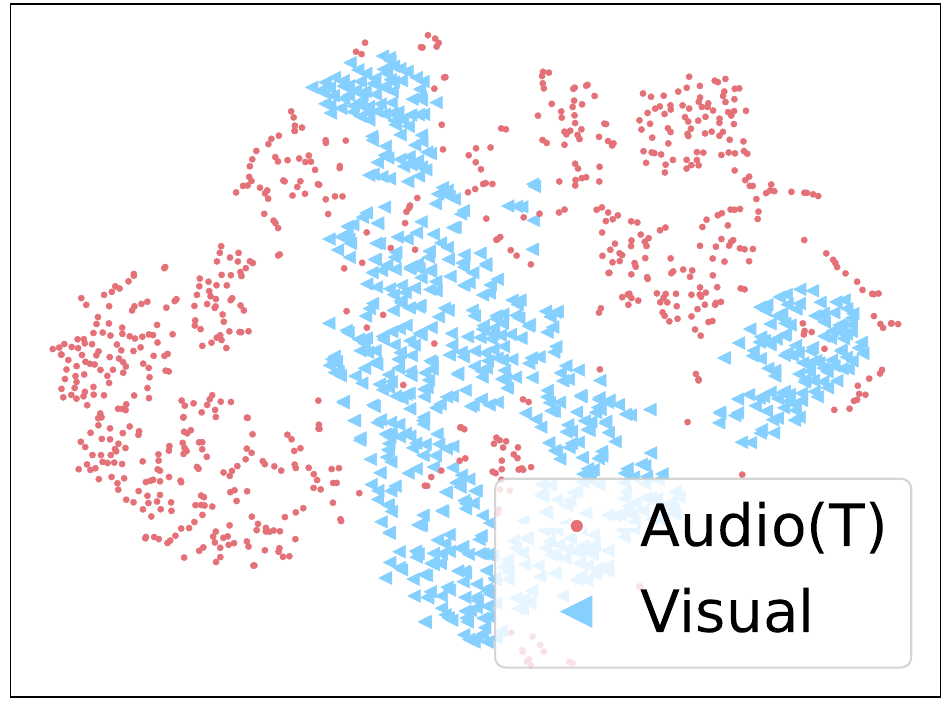}%
    \label{fig1:tsne1}}
    \hfil
    \subfloat[Ours]{\includegraphics[width=.33\columnwidth]{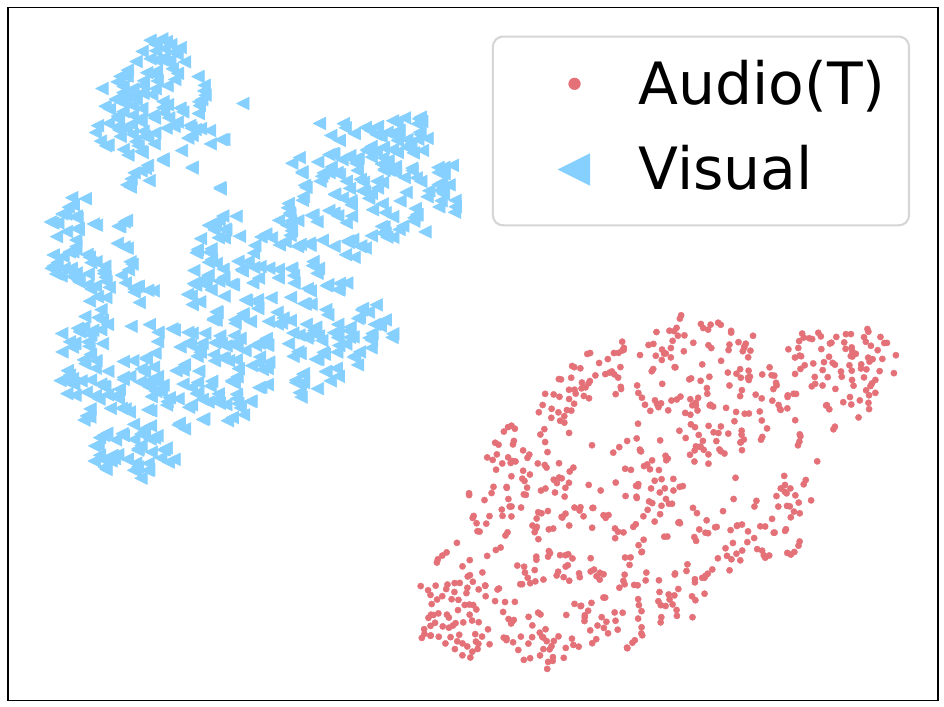}%
    \label{fig1:tsne2}}
    \caption{t-SNE visualization comparison between the conventional feature distillation method and our proposed approach. We visualize the features of different modalities on the CREMAD test set. T represents the teacher modality.}
    \label{fig:tsne}
\end{figure}

\subsection{Visualization}
In Figure \ref{fig:tsne}, we present a t-SNE \citep{van2008visualizing} visualization to compare the performance of the original feature distillation method and our proposed approach in cross-modal knowledge distillation. Figure \ref{fig1:raw} shows the result without any distillation, where the feature distributions of different modalities are clearly distinguishable. Figure \ref{fig1:tsne1} illustrates the result of traditional feature distillation, where there is significant overlap between the features of the visual modality (teacher) and audio modality (student). This indicates that the method fails to effectively differentiate between modality-specific features, leading to insufficient retention of modality-specific information, disrupting the original distribution of the student’s features, and reducing the student model's discriminative ability. In contrast, Figure \ref{fig1:tsne2} demonstrates the visualization of our approach, where the features from different modalities are clearly separated, forming two distinct clusters. This separation suggests that our method successfully disentangles modality-generic and modality-specific information, improving feature discrimination. These results validate our frequency decomposition strategy, which preserves modality-specific characteristics while enhancing cross-modal knowledge transfer.

\section{Conclusion}
In this paper, we investigate the non-negligible challenges faced by cross-modal knowledge distillation, particularly focusing on the discrepancies between modality-specific and modality-generic information, and the differences in feature distributions across modalities. Based on the observation and analysis of the failure of feature distillation in cross-modal scenarios, we propose a novel distillation framework. It decouples these types of information through frequency-based feature analysis and introduces a differentiated distillation strategy for different frequency components. Additionally, we address feature distribution discrepancies by incorporating a scale consistency loss and using a shared classifier for feature space alignment. Comprehensive experiments demonstrate the effectiveness of our approach.

\section{Acknowledgments} 
We would very much like to thank the anonymous reviewers for their valuable  comments. This work is supported in part by National Key
R\&D Program of China (Grant No. 2024YFB4505901), in part by the National Natural Science Foundation of China (Grant No. 62402024), in part by the Beijing Natural Science Foundation (Grant No. L241050), and in part by the Fundamental Research Funds for the Central Universities. For any correspondence, please refer to Dr. Renyu Yang (renyuyang@buaa.edu.cn). 


\bibliography{main}


\newpage

\section{EXPERIMENTAL SETUP}
\label{ap:1}
\subsection{Dataset}
\textbf{CREMA-D} \citep{cao2014crema} is a dataset for emotion recognition research, including two modalities of audio and vision. The dataset includes six emotion categories: happy, sad, angry, fear, disgust, and neutral. It contains 7,442 video clips, of which 6,698 are used as the training set and 744 are used as the test set.

\noindent \textbf{AVE} \citep{tian2018audio} is a dataset for audio-visual event localization, including two modalities of audio and vision. It contains a total of 4,143 videos, covering 28 event categories. The division of the training set, validation set, and test set refers to \citep{tian2018audio}.

\noindent \textbf{VGGSound} \citep{chen2020vggsound} is a large-scale audio-visual dataset consisting of more than 210,000 10-second videos. We randomly selected a dataset composed of 50 categories for experiments. In total, it includes 32,496 videos, of which 29,999 are divided into the training set and 2,497 are divided into the test set.

\noindent \textbf{CrisisMMD} \citep{alam2018crisismmd} is a multimodal dataset for research related to natural disasters. It consists of manually annotated tweets and pictures from Twitter, including two modalities of image and text. In total, it includes five categories (rescue, not humanitarian, affected individuals, infrastructure and utility damage, other relevant information). The division of the training set, validation set, and test set is the same as the original.

\noindent \textbf{NYU-Depth V2} \citep{icra_2019_fastdepth} is a multimodal dataset for indoor scene understanding research. It provides two modalities of depth information and RGB image information. There are a total of 40 categories. It contains 1,449 densely labeled RGB and depth image alignment pairs. Among them, 795 are divided into the training set and 654 are divided into the test set.

\subsection{Data preprocessing details}
We follow \citep{cao2025fastdrivevla,fan2024detached,wei2024enhancing} and provide our preprocessing details. For the audio-visual datasets, the audio data is converted into spectrograms with a size of 257×299 for CREMA-D and 257×1,004 for both AVE and VGGSound. The spectrograms are generated using a window length of 512 and an overlap of 353. For visual modality, during training, 1 frame is extracted from AVE and CREMA-D, while 3 frames are uniformly sampled from VGGSound. During testing, the middle frame is selected. Random cropping and flipping data augmentation methods are applied during training, and the same approach was used for the visual modality in the CrisisMMD dataset. For the NYU V2 dataset, we apply random HSV and random flipping as data augmentation techniques on the RGB modality.

\subsection{Network architectures}
For the audio-visual datasets, we use ResNet-18 as the backbone network. In the CrisisMMD dataset, we use the pre-trained BERT-base as the backbone for text and MobileNetV2 as the backbone for images. For the NYU V2 dataset, we use DeeplabV3+ with a ResNet-18 backbone as the network architecture. Additionally, during distillation, we use the features extracted before the ReLU activation. Since the core of segmentation tasks is to generate pixel-level classification results rather than mapping global features to a fixed class, our method does not use a shared classifier alignment module for segmentation tasks.

\subsection{Training details}
\textbf{Optimizer:} For BERT, we use the Adam optimizer, while for the others, we use the SGD optimizer with a momentum of 0.9.

\noindent \textbf{Learning rate:} For BERT, a fixed learning rate of 1e-5 is used. For segmentation tasks, the initial learning rate is 0.02 and decays according to the `poly' policy with a power of 0.9. For all other tasks, the initial learning rate is set to 1e-2 and follows the `poly' decay policy with a power of 0.9.

\noindent \textbf{Batch size:} For segmentation tasks, the batch size is 16, while for all other tasks, it is set to 64.

\noindent \textbf{Epochs:} For BERT, since it is pre-trained, we only train for 20 epochs. For segmentation tasks, we train for 50 epochs, and for all other tasks, we train for 100 epochs.

\subsection{Training environment}
All experiments are conducted on NVIDIA Tesla V100 GPUs using CUDA 11.8 with the PyTorch framework.

\subsection{Cross-Modal feature similarity}
Cross-Modal feature similarity Raw features (from trained unimodal models) decomposed to high/low frequencies. CM feature similarity is cosine based.

\begin{figure*}[t]
    \centering
    \includegraphics[width=.9\textwidth]{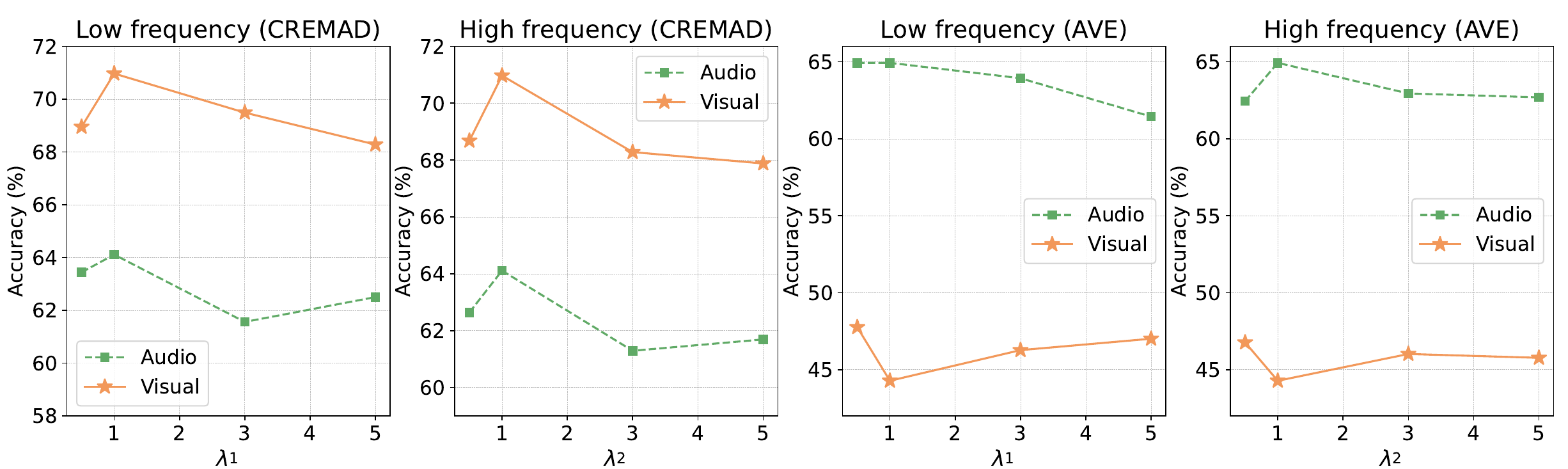}
    \caption{\textbf{Sensitivity study of high and low-frequency loss weight.}}
    \label{fig:weight}
\end{figure*}

\section{EXPERIMENTAL RESULT}
\subsection{More comparision on NYUv2}
\renewcommand{\multirowsetup}{\centering}

\begin{table}[t]
    \centering
    \renewcommand{\arraystretch}{1.1}
    \small
    \centering
    
    \begin{tabular}{p{2cm}| p{1.5cm} | p{1.5cm} }
        \toprule
        \textbf{Method} & \makecell[c]{\textbf{Depth}} & \makecell[c]{\textbf{RGB}}  \\

        \hline
        MMANet & \makecell[c]{29.6} & \makecell[c]{32.6} \\

        RDP & \makecell[c]{29.7} & \makecell[c]{27.2}  \\

        CIRKDv2 & \makecell[c]{33.1} & \makecell[c]{36.4}  \\

        Ours & \makecell[c]{33.2} & \makecell[c]{36.9}  \\

        CIRKDv2+Ours & \makecell[c]{35.1} & \makecell[c]{37.9}  \\

        \bottomrule
    \end{tabular}
    \caption{\textbf{The comparison on the semantic segmentation task.} The metric denotes the mean Intersection over Union.}
    \label{tab:more_comp}
\end{table}

We also compared some other methods on the segmentation task, including CIRKDv2~\cite{yang2023online} methods optimized for the segmentation task, as shown in Table \ref{tab:more_comp}. Our method achieved the optimal performance, and our method can further combine with CIRKDv2 to further enhance its performance.

\subsection{Abalation study of loss functions}
\renewcommand{\multirowsetup}{\centering}

\begin{table}[t]
    \centering
    \setlength{\tabcolsep}{12pt}
    \renewcommand\arraystretch{1.2}

    \small
    \centering
    \footnotesize 
    
    \begin{tabular}{p{1.5cm}| p{1.2cm}| p{0.8cm} | p{0.8cm}}
        \toprule
        \multicolumn{2}{c|}{\textbf{Loss}} & \multicolumn{2}{c}{\textbf{CREMAD}} \\
        \hline
        \makecell[c]{Low} & \makecell[c]{High} & \makecell[c]{A} & \makecell[c]{V} \\ 

        \hline
        \makecell[c]{MSE} & \makecell[c]{LogMSE} & \makecell[c]{64.1} & \makecell[c]{71.0}  \\

        \makecell[c]{MSE} & \makecell[c]{MSE} & \makecell[c]{62.2} & \makecell[c]{70.5}  \\

        \makecell[c]{LogMSE} & \makecell[c]{LogMSE} & \makecell[c]{62.6} & \makecell[c]{68.0}  \\

        \makecell[c]{LogMSE} & \makecell[c]{MSE} & \makecell[c]{61.7} & \makecell[c]{67.6}  \\



        \bottomrule
    \end{tabular}
    \caption{\textbf{The analysis of using different loss functions for different features.}}
    \label{tab:r1}
\end{table}
In this experiment, we compared different loss functions for low-frequency and high-frequency features, as shown in Table \ref{tab:r1}. The results indicate that the model achieves the best performance when MSE loss is used for low-frequency features and LogMSE loss is used for high-frequency features. By comparing the results of the first and third rows, as well as the second and fourth rows, we observe that MSE loss is most suitable for low-frequency features. Additionally, comparing the first and second rows, as well as the third and fourth rows, shows that LogMSE loss is most effective for high-frequency features. This further suggests that high-frequency features likely contain more modality-specific information, making strong alignment less suitable, whereas low-frequency features may contain more modality-generic information, making strong consistency more appropriate.

\subsection{Ablation study on frequency threshold}
\renewcommand{\multirowsetup}{\centering}

\begin{table}[t]
    \centering
    \setlength{\tabcolsep}{18pt}
    \renewcommand{\arraystretch}{1.1}
    \small
    \centering
    
    \begin{tabular}{c| p{1cm} | p{1cm} }
        \toprule
        \textbf{Threshold} & \makecell[c]{\textbf{A}} & \makecell[c]{\textbf{V}}  \\

        \hline
        1/4 & \makecell[c]{62.9} & \makecell[c]{70.7} \\

        1/2 & \makecell[c]{64.1} & \makecell[c]{71.0}  \\

        1/3 & \makecell[c]{62.1} & \makecell[c]{69.4}  \\

        \bottomrule
    \end{tabular}
    \caption{\textbf{Ablation study on frequency threshold.}}
    \label{tab:freq_thresh}
\end{table}

To investigate the effects of varying division thresholds, we performed ablation studies, with results illustrated in Table~\ref{tab:freq_thresh} (using the CREMAD dataset as a representative case). The findings reveal that employing a threshold of 1/2 yields the superior performance of the model, validating its efficacy in optimizing the partitioning process. Furthermore, future endeavors will delve into adaptive and learnable division thresholds, enabling even more versatile and dynamic separations tailored to diverse scenarios.

\subsection{Sensitivity study of frequency}
\renewcommand{\multirowsetup}{\centering}

\begin{table}[t]
    \centering
    \renewcommand{\arraystretch}{1}
    \small
    \centering
    
    \begin{tabular}{p{1.cm}| p{1.cm}| p{1cm} | p{1cm} |p{1cm} | p{1cm}}
        \toprule
        \multicolumn{2}{c|}{\textbf{Method}} & \multicolumn{2}{c|}{\textbf{CREMAD}} & \multicolumn{2}{c}{\textbf{AVE}} \\
        \hline
        \makecell[c]{High} & \makecell[c]{Low} & \makecell[c]{A} & \makecell[c]{V} & \makecell[c]{A} & \makecell[c]{V} \\ 

        \hline
        \makecell[c]{$\checkmark$} & ~ & \makecell[c]{62.4} & \makecell[c]{67.5} & \makecell[c]{64.9} & \makecell[c]{44.0} \\

        ~ & \makecell[c]{$\checkmark$} & \makecell[c]{62.1} & \makecell[c]{69.1} & \makecell[c]{64.2} & \makecell[c]{44.5} \\

        \makecell[c]{$\checkmark$} & \makecell[c]{$\checkmark$} & \makecell[c]{64.1} & \makecell[c]{71.0} & \makecell[c]{64.9} & \makecell[c]{47.8} \\

        \bottomrule
    \end{tabular}
    \caption{\textbf{The analysis of frequency loss.}}
    \label{tab:frequency}
\end{table}

We employ high-frequency and low-frequency features to imitate the teacher model, respectively. In this subsection, we conduct experiments on high-frequency loss and low-frequency loss to investigate their influences on performance. As shown in Table \ref{tab:frequency}, both the high-frequency loss and low-frequency loss lead to significant accuracy improvements. Furthermore, considering the frequency with character, we find high-frequency benefits more to the audio modality and low-frequency benefits more to the visual modality. Besides, when combining the high-frequency and low-frequency loss, we achieve the best performance, which indicates that transferring modality-generic information and modality-specific information can complement each other effectively to enhance the overall performance.

\subsection{Sensitivity study of loss coefficients}
We utilize the coefficients $\lambda_1$ and $\lambda_2$ to balance the low-frequency and high-frequency, respectively. In this subsection, we do the sensitivity study of weight loss on CREMA-D and AVE. The results are shown in Figure \ref{fig:weight}. We conduct experiments by setting the weights to 0.5, 1, 3, and 5, respectively. On the CREMA-D dataset, both modalities achieve the best performance when $\lambda_1 = 1$ and $\lambda_2 = 1$. On the AVE dataset, the audio modality performs best with $\lambda_1 = 1$ and $\lambda_2 = 1$, while the visual modality achieves the highest accuracy with $\lambda_1 = 0.5$ and $\lambda_2 = 0.5$. Therefore, $\lambda_1 = 1$ and $\lambda_2 = 1$ generally represent a good configuration.



\end{document}